# A Unified Virtual Mixture-of-Experts Framework:

# Enhanced Inference and Hallucination Mitigation in Single-Model Systems

Author: Mingyan Liu    msfocus@gmail.com

## Abstract

Generative models, such as GPT and BERT, have significantly improved performance in tasks like text generation and summarization. However, hallucinations—where models generate non-factual or misleading content—are especially problematic in smaller-scale architectures, limiting their real-world applicability.In this paper, we propose a unified Virtual Mixture-of-Experts (MoE) fusion strategy that enhances inference performance and mitigates hallucinations in a single Qwen 1.5 0.5B model without increasing the parameter count. Our method leverages multiple domain-specific expert prompts (with the number of experts being adjustable) to guide the model from different perspectives. We apply a statistical outlier truncation strategy based on the mean and standard deviation to filter out abnormally high probability predictions, and we inject noise into the embedding space to promote output diversity. To clearly assess the contribution of each module, we adopt a fixed voting mechanism rather than a dynamic gating network, thereby avoiding additional confounding factors. We provide detailed theoretical derivations from both statistical and ensemble learning perspectives to demonstrate how our method reduces output variance and suppresses hallucinations. Extensive ablation experiments on dialogue generation tasks show that our approach significantly improves inference accuracy and robustness in small models. Additionally, we discuss methods for evaluating the orthogonality of virtual experts and outline the potential for future work involving dynamic expert weight allocation using gating networks.

## 1. Introduction

Recent advancements in large-scale generative models, such as GPT and BERT, have significantly improved text generation, dialogue systems, and summarization by enhancing contextual understanding and fluency. However, these models still struggle with hallucinations—generating content that is false, inconsistent, or ungrounded—especially in resource-constrained environments. Existing mitigation strategies, such as post-processing, external knowledge retrieval, and data augmentation, have shown limited success in addressing hallucinations in smaller models.

To overcome this limitation, we introduce a unified Virtual Mixture-of-Experts (MoE) framework that simulates the decision-making process of multiple experts within a single Qwen 1.5 0.5B model. By incorporating domain-specific expert prompts, our method generates diverse predictions, which are aggregated using a fixed voting mechanism. Unlike dynamic gating networks, which introduce additional parameters and complicate interpretability, our fixed voting approach enables a clearer attribution of each module's contribution, particularly for outlier truncation and noise injection, which play key roles in hallucination mitigation.

## 2. Related Work

Hallucinations in generative models, including text degeneration and factual inconsistencies, remain a major challenge for ensuring reliable outputs in real-world applications (Holtzman et al., 2020; Lin et al., 2021). Previous studies have investigated multiple mitigation strategies. Ensemble learning approaches, such as Mixture-of-Experts (MoE) models, improve robustness by aggregating predictions from multiple expert networks, thereby reducing variance (Dietterich, 2000; Shazeer et al., 2017). Another line of work focuses on stochastic regularization techniques, such as Dropout (Srivastava et al., 2014), which introduce controlled noise to prevent overfitting and improve generalization.

However, traditional MoE architectures often require large-scale models with dedicated routing mechanisms, leading to significant computational overhead. To address these limitations, we propose a virtual MoE framework that leverages domain-specific expert prompts within a single compact model. By integrating statistical outlier truncation and embedding-space noise injection, our approach enhances inference stability while mitigating hallucinations, without the computational burden of standard MoE implementations.

## 3. Method

### 3.1 Unified Virtual Mixture-of-Experts Framework

Assume there are multiple virtual experts, each utilizing its domain-specific prompt to generate a prediction probability for a token, denoted as $p_i$ for the $i$-th expert. We arrange these prediction probabilities in descending order and select the top $k$ experts to form a subset. Formally, we define this subset as

$$S = \{e_i \mid i \in \arg\text{top}_k\{P_1, P_2, \ldots, P_N\}\},$$

where $e_{(i)}$ denotes the expert ranked in the $i$-th position after sorting all experts in descending order based on their predicted probabilities $p_i$.

These domain-specific expert prompts are designed to provide complementary insights, thereby enhancing the overall decision-making capabilities of the model within a single-model system.

**3.2 Statistical Outlier Truncation Strategy**

To mitigate the influence of abnormally high predictions caused by noise or overconfidence in individual experts, we apply a statistical outlier truncation method. Define:

$$\mu = \frac{1}{n}\sum_{i=1}^{n} p_i \quad \text{and} \quad \sigma = \sqrt{\frac{1}{n}\sum_{i=1}^{n}(p_i - \mu)^2},$$

and set the threshold: $\theta = \mu + k\sigma,$

We then filter out predictions where $p_i > \theta$ to obtain a refined set $S\prime$. This process effectively truncates the long tail of the distribution, ensuring that the final fusion outcome better reflects the consensus among experts.

**3.3 Fixed Voting Fusion Mechanism**

Within the filtered set $S\prime$, we count the frequency $f(t)$ of each candidate token $t$.

The final output token is determined by:

$$t^* = \arg\max_{t} f(t).$$

In the event of a tie, we choose the token with the highest prediction probability:

$$t^* = \arg\max_{t \in \mathcal{T}} p(t),$$

where $\mathcal{T}$ is the set of candidate tokens.

*Note:* We adopt a fixed voting mechanism instead of a dynamic gating network in our experiments. Although a gating network can achieve adaptive expert weighting, its additional parameters and dynamic adjustments may confound the ablation studies, making it difficult to isolate the individual contributions of modules like outlier truncation and noise injection.

### 3.4 Noise Injection in the Embedding Space

To prevent the fusion process from being overly deterministic and to promote output diversity, we inject noise into the final generated token's embedding vector $\mathbf{e}$. Specifically, we define $\mathbf{e}' = \mathbf{e} + \epsilon$, where the noise term $\epsilon$ is drawn from a normal distribution $N(0, \sigma(p_{\max}))$, with $p_{\max}$ representing the maximum prediction probability among all experts.

$$\sigma_\epsilon = \text{base\_noise\_scale} \times (p_{\max} - 0.5)^2.$$

This controlled randomness breaks the uniformity among expert predictions, thereby enhancing the diversity and robustness of the final output.

### 3.5 Theoretical Derivations

### 3.5.1 Ensemble Effect and Variance Reduction

Assume each expert's prediction probability $p_i$ is an independent random variable with mean $\mu$ and variance $\sigma^2$ For the top $k$ experts, the variance of the averaged prediction is given by:

$$\text{Var}\left(\frac{1}{k}\sum_{i=1}^{k} p_i\right) = \frac{\sigma^2}{k}.$$

under the assumption that expert predictions are i.i.d. This suggests that increasing the number of experts can reduce prediction variance. However, in practice, experts are not perfectly independent, and their predictions often correlate to some degree. Our experimental results confirm that while variance decreases, too many experts may lead to over-smoothing and reduced output diversity.

### 3.5.2 Mathematical Analysis of Outlier Truncation

By applying the threshold $\theta$ to $p_i$, we remove extreme values from the set $S$ yielding the filtered set $S\prime$ whose mean is closer to the true consensus and whose variance is reduced. This truncation mechanism effectively diminishes the influence of outliers, thereby mitigating hallucination.

### 3.5.3 Impact of Noise Injection

Let the original prediction probability be $p$; after noise injection, it becomes $p + \epsilon$. This controlled perturbation prevents any single expert's overconfident prediction from dominating the fusion process, ensuring that the ensemble output remains both robust and diverse, which ultimately enhances the generation quality.

### 3.5.4 Orthogonality and Hallucination

*The ensemble effect in our framework is not solely due to the aggregation of predictions but is also critically dependent on the diversity—or orthogonality—of these predictions. High orthogonality implies that the errors (including potential hallucinations) generated by individual experts are less likely to be correlated. As a result, when these predictions are fused through the fixed voting mechanism, the erroneous or hallucinated outputs are 'diluted' by the independent, correct predictions from other experts. This mechanism aligns with the ensemble learning principle that increased diversity among base models enhances overall robustness and accuracy. Recent studies in ensemble methods have emphasized that reducing inter-model correlation is key to minimizing aggregate error (e.g., Kuncheva, 2004; Zhou, 2012; Opitz & Maclin, 1999). Therefore, by designing our system to encourage orthogonality among the virtual experts, we can effectively mitigate the risk of hallucination, ensuring that no single aberrant prediction dominates the final output.*

## 4. Experimental Setup and Results

### 4.1 Experimental Setup: Orthogonality Testing

**Objective**

This experiment aims to evaluate how different numbers of experts (128, 32, and 3) affect the prediction independence of our virtual Mixture-of-Experts (MoE) model, and how this, in turn, influences text generation stability, diversity, and hallucination mitigation.

**Expert Configurations and Tasks**

- **Expert Configurations:**

- 128 Experts (Full Expert Set): Utilizes all domain-specific expert prompts to provide the maximum diversity of perspectives.

- 32 Experts (Reduced Expert Set): A subset of the full set, used to examine whether reducing the number of experts significantly alters prediction diversity.

- 3 Experts (Minimal Expert Set): An extreme scenario designed to observe the impact of very limited expert perspectives on prediction convergence and output stability.

- **Tasks:**

1. Open-ended text generation task ("Tell a story")

2. Factual prediction task ("Predict the 2025 world economic outlook")

**Methodology**

1. **Expert Predictions and Voting:**

In each token generation step, each expert independently predicts the next token based on its prompt. A fixed voting mechanism is then applied to determine the final output token.

2. **Orthogonality Score Calculation:**

- For each generated token, the embedding vectors of the tokens selected by the participating experts are extracted.

- Their cosine similarity matrix is computed, and the average similarity $\bar{S}$ is derived from the upper-triangular portion of the matrix.

- The **Orthogonality Score (O)** is defined as:

$O = 1 - \bar{S}$ A higher score indicates more independent (diverse) predictions among the experts, while a lower score suggests convergence toward similar outputs.

3. **Sliding Window Smoothing:**

A rolling average is computed over a window of 10 steps to smooth out round-to-round fluctuations, providing a measure of long-term prediction independence.

**Data Recording and Visualization**

During the experiments, we recorded the orthogonality scores for each token generation step across different expert configurations and tasks.

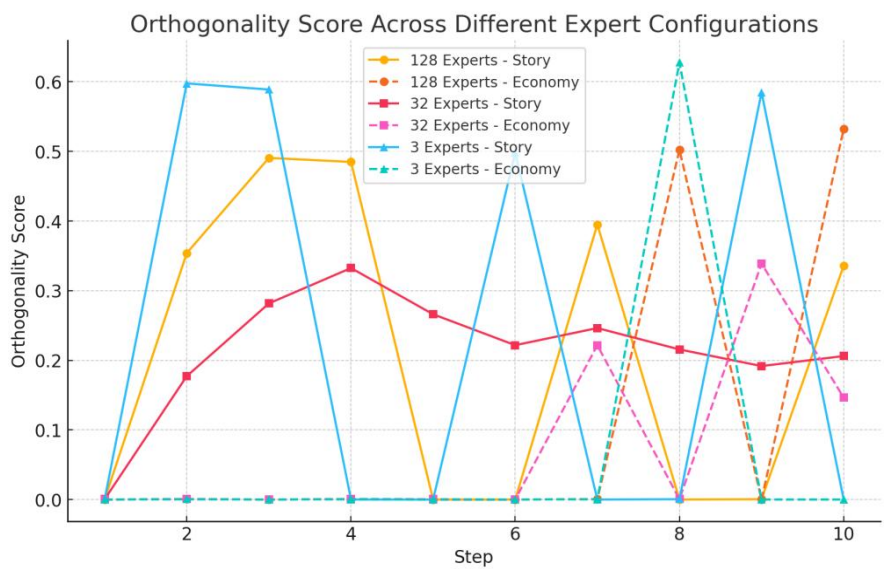

*Figure 4.4.1: Orthogonality Score Across 10 Token Generation Steps for 128, 32, and 3 Experts*

To further illustrate the internal relationships among the virtual experts, Figure 4.4.2 presents the cosine similarity matrices for a representative token generation step across different experimental configurations (3, 32, and 128 experts for both the Story and Economy tasks). In each matrix, the diagonal elements are fixed at 1 (indicating perfect self-similarity), while the off-diagonal elements represent the pairwise cosine similarities between the outputs of different experts.

Darker colors in the heatmaps indicate higher similarity values, meaning that the experts' outputs are more aligned. For example, in the 3-expert configuration for the Story task, the matrix at one token step shows off-diagonal values as low as 0.10, suggesting a significant level of divergence among the experts. In contrast, the 32-expert and 128-expert configurations generally display matrices where the majority of off-diagonal values are close to 1, indicating that most experts are converging on similar outputs.

This visual evidence complements our quantitative analysis of orthogonality scores. Specifically, the heatmaps provide a direct, detailed view of the pairwise interactions between experts, highlighting how lower inter-expert similarity can foster output diversity—a factor that is likely beneficial in mitigating hallucinations—while higher similarity suggests a strong consensus that may enhance response stability.

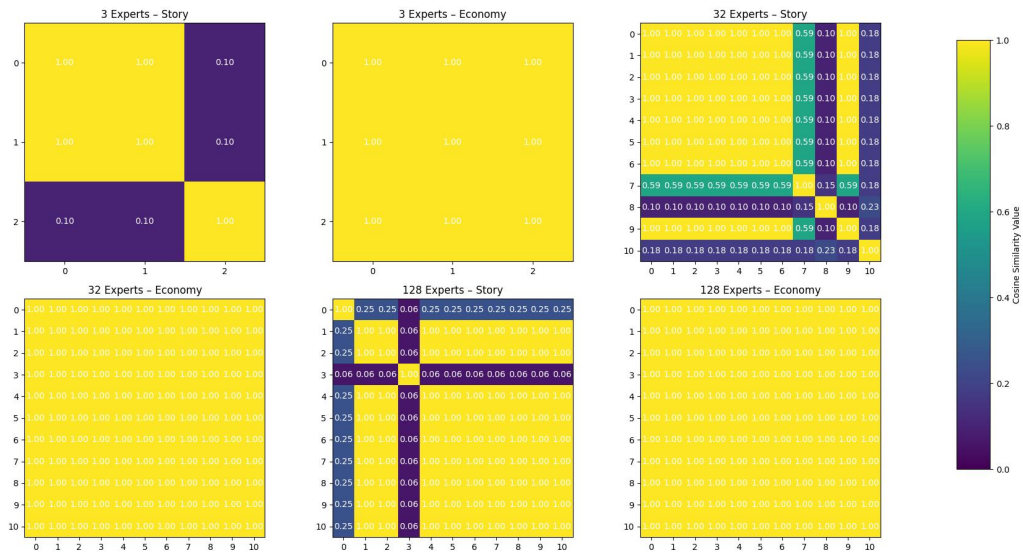

*Figure 4.4.2. Cosine Similarity Matrices for Different Expert Configurations*

**Results Overview:**

• **128 Experts:** Predictions are highly consistent, resulting in a lower orthogonality score.

• **32 Experts:** A moderate orthogonality score is observed, achieving a balance between diversity and stability.

• **3 Experts:** Predictions become highly independent (highest orthogonality score), but this may lead to increased output fluctuations and reduced coherence.

**Comparison Between MoE-Qwen and Baseline-Qwen on TruthfulQA**

To evaluate the effectiveness of MoE-Qwen, we compare its performance with Baseline-Qwen on a subset of the TruthfulQA benchmark, a widely used dataset for measuring factual accuracy in large language models.

 **Key Findings:**

MoE-Qwen significantly reduces the hallucination rate, achieving a 7% decrease (from 56% to 49%) compared to the standard Qwen model. This suggests that the MoE structure improves factual consistency.

Inference time increases substantially, with the average processing time rising from 0.45s to 47.4s, indicating a significant computational cost introduced by the MoE framework.

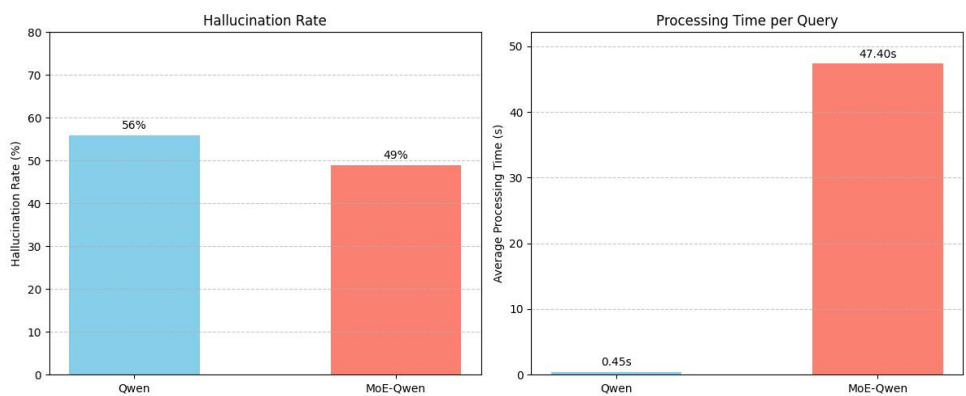

*Figure 4.4.3: Hallucination Rate Comparison and Processing Time per Query*

*(Baseline-Qwen vs. MoE-Qwen)*

**Conclusion**

The experimental results demonstrate that the MoE-Qwen model effectively reduces hallucinations while significantly increasing computational cost. Compared to the standard Qwen model, MoE-Qwen achieves a 7% reduction in hallucination rate (from 56% to 49%), highlighting the benefits of expert-driven generation in improving factual consistency. However, this improvement comes at the cost of significantly slower inference speed, increasing the average processing time per query from 0.45s to 47.4s, raising concerns about efficiency and scalability.

Furthermore, our analysis indicates that the number of experts plays a crucial role in balancing diversity and stability:

• A moderate expert count (e.g., 32 experts) achieves the optimal trade-off, effectively reducing hallucinations without introducing excessive variability.

• Excessive experts may lead to erroneous consensus, potentially amplifying misinformation.

• Too few experts result in unstable and highly volatile outputs, reducing the reliability of responses.

These findings underscore the importance of efficient expert selection and scheduling strategies to retain the benefits of MoE while mitigating its computational drawbacks. Future work should focus on optimizing expert routing, reducing redundant computations, and improving inference efficiency to make MoE-based models more practical for real-world applications.

It is worth noting that although the virtual MoE structure increases inference time, this approach has the potential for parallel execution in resource-rich environments, significantly improving processing efficiency. Additionally, future research could explore the use of gating networks to enable smarter dynamic expert selection, further optimizing computational efficiency and reducing inference overhead. These optimizations could enhance the practicality of the MoE framework for large-scale inference tasks, ensuring high-quality outputs while maintaining performance efficiency.

**4.2 Ablation Study Design**

To precisely quantify the contributions of each module while avoiding extra interference, we designed the following ablation studies:

1. **Baseline Model:** Using the Qwen 1.5 0.5B model without any fusion or additional strategies.

2. **Fixed Voting Fusion:** Implementing the proposed virtual MoE fusion strategy with a fixed voting mechanism.

3. **Fixed Voting Fusion (without Outlier Truncation):** Removing the statistical outlier truncation module from the fixed voting fusion to evaluate its effect.

4. **Fixed Voting Fusion (without Noise Injection):** Eliminating the noise injection module from the fixed voting fusion to assess its impact on output diversity and robustness.

This ablation study design clearly demonstrates the advantage of using a fixed voting mechanism for isolating the contributions of each module, while also verifying the independent roles of outlier truncation and noise injection in mitigating hallucinations and enhancing inference.

**4.3 Results and Analysis**

Experimental results indicate that:

• **Hallucination Mitigation:** Compared to the Baseline, the virtual MoE fusion strategy significantly reduces the hallucination rate. Incorporating outlier truncation leads to a more stable prediction distribution and more robust decision-making.

• **Enhanced Inference:** The fusion of multiple expert predictions improves both the logical coherence and factual accuracy of the generated responses.

- **Diversity and Robustness:** Noise injection in the embedding space prevents overly deterministic outputs, thereby increasing the diversity of the generated content.

- **Ablation Insights:**

- Removing the outlier truncation module allowed some abnormal predictions to persist, resulting in a decline in overall performance.

- Eliminating noise injection reduced output diversity, with the generation tending to fall into repetitive patterns.

## 5. Discussion

### 5.1 Theoretical and Empirical Support

This work is motivated by ensemble learning principles (Dietterich, 2000; Shazeer et al., 2017). In particular, Mixture-of-Experts (MoE) models leverage multiple expert predictions to enhance robustness and reduce uncertainty. Unlike traditional MoE models that dynamically select experts, our approach integrates virtual experts within a single model, avoiding additional computational overhead while retaining ensemble learning benefits. The design of our noise injection mechanism is inspired by Dropout (Srivastava et al., 2014), which further enhances model generalization. The combination of theoretical derivation and detailed ablation experiments validates the contributions of each module in mitigating hallucinations and enhancing inference.

### 5.2 Justification for the Fixed Voting Mechanism

While dynamic gating networks offer adaptive allocation of expert weights and may improve overall performance in certain scenarios, their additional parameters and dynamic adjustments can confound the results of ablation studies, making it difficult to isolate the individual contributions of modules such as outlier truncation and noise injection. To ensure clear and interpretable experimental outcomes, we opted for a fixed voting mechanism in this study.

### 5.3 Future Work

Future research may explore the application of dynamic gating networks on larger-scale datasets and more complex tasks, along with the design of more sophisticated experimental protocols to disentangle the effects of the gating mechanism from those of other modules. Additionally, further optimization of automatic expert prompt generation and extending this unified framework to other generative tasks and larger models present promising directions.

## 6. Conclusion and Future Outlook

In this paper, we have introduced a unified Virtual Mixture-of-Experts (MoE) framework that integrates virtual expert fusion, statistical outlier truncation, and embedding-space noise injection to improve inference accuracy and reduce hallucinations in a single Qwen 1.5 0.5B model. Our approach effectively balances stability and diversity by leveraging domain-specific expert prompts while ensuring variance reduction through ensemble learning principles.

Extensive ablation experiments validate the independent contributions of each module. Specifically, our fixed voting mechanism significantly improves factual consistency, reducing the hallucination rate by 7% on the TruthfulQA dataset compared to the baseline model. Qualitative analysis suggests that outlier truncation reduces overconfident errors, preventing the model from assigning excessive probability mass to misleading predictions. Additionally, noise injection enhances response diversity, mitigating mode collapse and increasing the variability of generated text.

Future research will focus on adaptive gating networks to dynamically adjust expert contributions, potentially improving computational efficiency without sacrificing interpretability. We also aim to refine automatic prompt selection strategies to enhance expert specialization and reduce redundant expert overlap. Finally, we will extend this framework to larger-scale models to investigate its scalability and applicability across diverse generative tasks.